\documentclass{article}
\usepackage{ijcai23}
\usepackage{times}
\usepackage{soul}
\usepackage{url}
\usepackage[hidelinks]{hyperref}
\usepackage[utf8]{inputenc}
\usepackage[small]{caption}
\usepackage{graphicx}
\usepackage{float} 
\usepackage{subfigure} 
\usepackage{mathtools}
\usepackage{amsmath}
\usepackage{amsthm}
\usepackage{booktabs}
\usepackage{algorithm}
\usepackage{algorithmic}
\usepackage[switch]{lineno}

\usepackage{bbold}

\title{RePaint-NeRF: NeRF Editting via Semantic Masks and Diffusion Models}

\author{
Xingchen Zhou$^{1,2}$
\and
Ying He$^{1,2}$\and
F Richard Yu$^{1,2}$ \and
Jianqiang Li$^1$ \and
You Li$^2$
\affiliations
$^1$Shenzhen University\\
$^2$Guangdong Laboratory of Artificial Intelligence and Digital Economy (SZ) \\
\emails
zhouxingchen2021@email.szu.edu.cn,
\{heying, yufei, lijq\}@szu.edu.cn,
liyougis@gmail.com
}

\begin{document}

\maketitle

\begin{figure*}[htbp] \centering \includegraphics[width=2\columnwidth,height=0.5\linewidth]{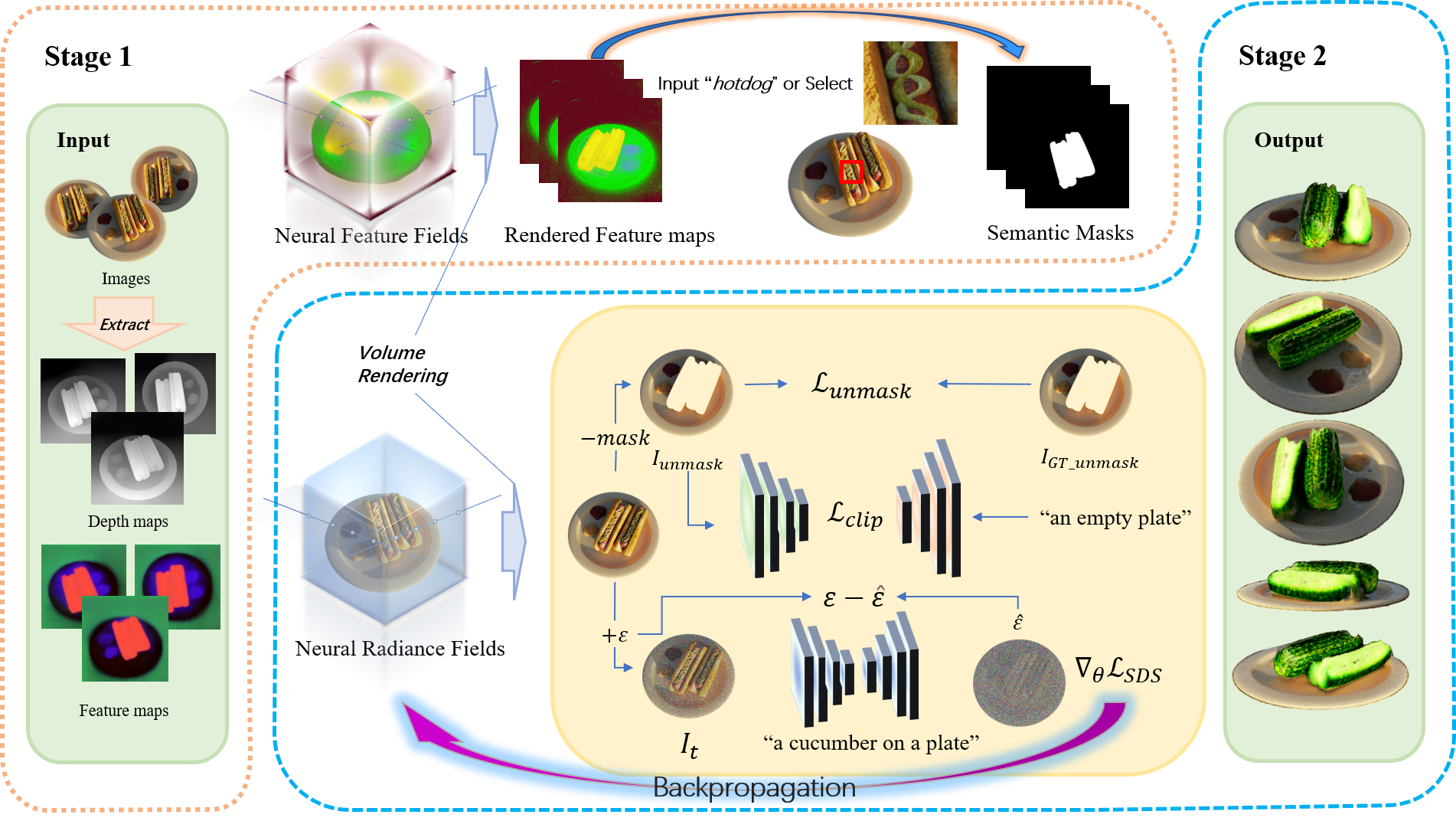}%
\caption{\textbf{Overview of RePaint-NeRF.} We present an editing method in NeRF. In the first stage, we additionally optimize a feature field along with the color module and density module 
to extract the content mask by using text or patch. In another way of speaking, we separate the part we want to change for a generation. Then, we use the mask and text prompt to generate the new content guided by the pre-trained diffusion model and CLIP model. After optimization of the generation, we can finally repaint a pre-trained NeRF model with view consistency and 
scene integrity.} 
\label{overview} 
\end{figure*}

\begin{abstract}
The emergence of Neural Radiance Fields (NeRF) has promoted the development of synthesized high-fidelity views of the intricate real world. However, it is still a very demanding task to repaint the content in NeRF. In this paper, we propose a novel framework that can take RGB images as input and alter the 3D content in neural scenes. Our work leverages existing diffusion models to guide changes in the designated 3D content. Specifically, we semantically select the target object and a pre-trained diffusion model will guide the NeRF model to generate new 3D objects, which can improve the editability, diversity, and application range of NeRF. Experiment results show that our algorithm is effective for editing 3D objects in NeRF under different text prompts, including editing appearance, shape, and more. We validate our method on both real-world datasets and synthetic-world datasets for these editing tasks. Please visit \url{https://starstesla.github.io/repaintnerf/} for a better view of our results.
\end{abstract}

\section{Introduction}
High-quality reconstruction of a complex 3D world is a critical challenge in computer vision~\cite{aharchi2020review}. Neural Radiance Fields (NeRF)~\cite{mildenhall2021nerf} is an advanced approach for reconstructing the photo-realistic view of real 3D scenes. Nevertheless, most NeRF models implicitly encode the 3D scene by multiple layer perceptions (MLP)~\cite{mildenhall2021nerf} or spherical harmonics~\cite{fridovich2022plenoxels}, the shape and appearance of the scene can only be seen after rendering, which means its content cannot be edited as we do in explicit scenarios. In the scenario of automatic driving, an automatic driving model requires a large amount of realistically simulated data for training~\cite{li2019aads}. NeRF can provide large volumes of high-fidelity data for self-driving training to alleviate the gap of Sim2Real~\cite{tancik2022block}. However, there is still a certain distance for the current NeRF to produce numerous simulated data among the implicit scene, which greatly limits the scope of the application of NeRF. Therefore, editing within NeRF is essential in many cases.

In prior research studies~\cite{kobayashi2022decomposing,yang2021learning,kundu2022panoptic} that decompose the implicitly encoded scene for editing the objects in the scene by assigning semantic labels to each point in three-dimensional space. Then they can edit the content in NeRF by manipulating the collection of labeled points. For example, by setting the density of a car on the street to zero, it can be removed from the scene, or a red car can be set to blue by modifying its RGB value. However, these operations cannot meet creative editing needs, such as turning a pickup truck into a sedan, which requires models with strong generalization capabilities to change the shape and appearance of objects in space.

Recent works~\cite{ramesh2022hierarchical,rombach2022high} have shown very promising results in generating image content through text prompts. Users are now free to edit 2D images using text prompts and generate new images at a higher resolution. One of the keys is the thousands of rich images on the Internet, which enables the model to understand the content in the image and align with the abstract concepts in the language. In the three-dimensional domain, the lack of diverse 3D data limits the development of such generative models~\cite{lin2022magic3d}.

A recent approach, DreamFusion~\cite{poole2022dreamfusion} integrates a 2D pre-trained diffusion model~\cite{saharia2022imagen} with NeRF to generate 3D objects from text prompts. In more detail, they use an optimized gradient from the denoising process of diffusion model~\cite{ho2020denoising} to update the NeRF model in the direction of the text prompt and finally obtain a 3D model that conforms to the description of the text prompt and ensures view-consistency. However, DreamFusion's high memory and time consumption limits its scalability, making it impractical for generating complex 3D scenes.

To tackle these problems, we propose a new framework for editing the content in NeRF from text prompts. More specifically, we first mask the area to be edited, under the guidance of the pre-trained text-to-image diffusion model~\cite{poole2022dreamfusion}, we can modify the specified area from text prompts. However, manually smearing the mask on the two-dimensional training images cannot guarantee view consistency, and consumes a lot of time. Thus, we split our framework into two stages. In the first stage, based on the vanilla NeRF~\cite{mildenhall2021nerf} encoding color and density, we additionally extend a semantic feature module to provide users with semantic target selection. These semantic features are extracted from a pre-trained large-scale model, such as CLIP~\cite{radford2021learning}. After the first stage of training, users can select a patch to get the target object mask and generate new training data with mask information. In the second stage, based on DreamFusion~\cite{poole2022dreamfusion}, we gradually modify the mask area to conform to the shape and appearance of the text prompts under the guidance of the diffusion model. Moreover, we find that if the newly generated object is smaller than the original object, the previously covered area will be exposed, but since this part of the content is unknown to the model, this area will become a black hole. To alleviate this problem, we additionally add a background prompt to guide the generation of the content of the black hole. However, we find that only adding a background prompt is not enough. Inspired by~\cite{mirzaei2022laterf}, we use CLIP~\cite{radford2021learning} to encourage the filling of this part. We name our method, RePaint-NeRF, which means that based on a pre-trained NeRF, we can recreate the content inside it. Our experiments on both real-world and synthetic datasets demonstrate the effectiveness of our method in changing the content in different scenes under various text prompts.

In summary, our contributions include:

\begin{itemize}
    \item We propose a new framework that is capable of editing 3D content in NeRF through text prompts. To the best of our knowledge, we are the first work to propose editing NeRF using a diffusion model in complex scenes.
    \item Our method can greatly expand the scope of the application of NeRF model and apply it to most existing NeRF model architectures.
    \item Our approach enables practical semantic-masked object editing, making it possible for guiding editing in continuous NeRF scenes by diffusion models.
\end{itemize}

\section{Related Work}
Our method mainly utilizes a pre-trained text-to-image diffusion model~\cite{rombach2022high} for NeRF editing. In this section, we mainly summarize some recent NeRF editing research and text-to-content generation works.

\subsection{Neural Radiance Fields Editing}

Neural Radiance Fields (NeRF)~\cite{mildenhall2021nerf} uses a multi-layer perceptual layers network to encode complex scenes in a coordinate system-based manner and render high-quality 3D views in an end-to-end manner. A large amount of variant works~\cite{barron2021mip,muller2022instant,fridovich2022plenoxels,pumarola2021d} were released, setting off a wave of neural rendering. However, most NeRF variant works are based on implicit neural representations, which makes NeRF not as easy to edit as traditional explicit primitives, such as mesh. Some NeRF editing studies~\cite{wang2022clip,liu2021editing,yuan2022nerf,xu2022deforming,kobayashi2022decomposing} recently are proposed to address this challenging issue.

\paragraph{Object-level NeRF Editing.}  Some of the NeRF editing works~\cite{wang2022clip,liu2021editing} focus on a single class of objects. For example, Editing-NeRF~\cite{liu2021editing} can change the shape or color of some parts of a certain class of objects, such as a chair or a car. Specifically, Editing-NeRF trains a neural network on a large number of a single category of objects, which is designed to learn the shape code and appearance code of these 3D models. Editing-NeRF can edit the shape and appearance of an object by adjusting these two latent codes. However, this method can only be operated on similar objects and can hardly extend the editing operation to complex scenes.

\paragraph{Neural Scene Decomposition.} Another research direction~\cite{kobayashi2022decomposing,kundu2022panoptic,zhi2021place} is to decompose the neural scene first and add semantic labels to each 3D coordinate point by additionally training a semantic branch so that a certain class of object can be selected to edit during rendering. However, such methods are limited in that they cannot extend editing to invisible content, such as turning a rock into an apple. Our work relies on a pre-trained diffusion model, which endows the power to regenerate a selected object in a complex scene.

\subsection{Text-to-3D Generation}
Recently, some methods~\cite{poole2022dreamfusion,lin2022magic3d,metzer2022latent,jain2022zero} have been proposed to transfer knowledge from pre-trained 2D diffusion models to 3D fields. DreamField~\cite{jain2022zero} uses a pre-trained CLIP~\cite{radford2021learning} model to supervise the gap between the views rendered from different perspectives represented by NeRF and a text prompt. However, the generated 3D models are still not photo-realistic. The recently proposed DreamFusion~\cite{poole2022dreamfusion} and its variants ~\cite{lin2022magic3d,metzer2022latent} use pre-trained diffusion models to guide the generation of 3D models. The diffusion model generates gradients through its denoising mechanism~\cite{nichol2021improved,ho2020denoising} by randomly looking at the 3D field, the gradient is then passed directly to the NeRF model for optimization. However, these approaches can not extend to scene-level generation due to memory limitations. In our paper, we use the diffusion model for the editing of different objects in the scene, achieving scene-level generation in a sense.

\section{Method}

The first part of our method is to mask the places we want to modify. However, it is too time-consuming to manually paint masks from different angles in a 3D scene. Thus, to get a view consistent semantic mask, we additionally train a semantic feature module to obtain a relative view continuous mask. In detail, we encode the 3D coordinate points to a high-dimension feature space. Moreover, we also add depth maps predicted from training images to supervise the depth of the predicted view, which could effectively reduce noise and speed up training~\cite{deng2022depth}. The second part of our framework is about how to generate a new object by text guidance based on the target mask. We also need to preserve the other existing content, so we add background prompts and a CLIP loss to monitor the plausibility of background content. We extract the pose information of the training views via COLMAP~\cite{schonberger2016structure}. For the supervised depth information, we use an existing robust model~\cite{Ranftl2022} which could predict a rough depth of a monocular image. Note that the depth estimation model can be replaced by other models that predict depth relatively accurately. For the supervised semantic features, we refer to DFF~\cite{kobayashi2022decomposing}, which distills the feature maps from a pre-trained CLIP model~\cite{li2022language}. The overview of our framework is shown in Fig.~\ref{overview}.
   
\subsection{Preliminaries}
\paragraph{Neural Radiance Fields.} NeRF~\cite{mildenhall2021nerf} implicitly encodes the color $\boldsymbol{c}$ and density $\sigma$ of each 3D point $\boldsymbol{x}$ from different view direction $\boldsymbol{d}$ by utilizing multiple perceptual layers $g_{\theta} :(\boldsymbol{x},\boldsymbol{d}) \to (\boldsymbol{c},\sigma)$ weighted by $\theta$. Considering a ray $\boldsymbol{r}(t) = \boldsymbol{o} + t\boldsymbol{d}$ is emitted to sample the points along the ray in 3D space, where $\boldsymbol{o}$ is the origin of the ray, $t$ is the distance from the origin $\boldsymbol{o}$ to the sample point $\boldsymbol{x}$ along the ray. The color of a pixel can be obtained by volume rendering:

\begin{equation}
    C(\boldsymbol{r})=\int_{t_{n}}^{t_{f}} T(t) \sigma(\boldsymbol{r}(t)) c(\boldsymbol{r}(t), \boldsymbol{d}) \mathrm{d}t,\\
\end{equation}
\begin{equation}
    \mathrm{where} \quad T(t)=\exp \left(-\int_{t_{n}}^{t} \sigma(\boldsymbol{r}(t)) \mathrm{d} t\right),
\end{equation}
where $T(t)$ can be regarded as transparency, $t_n$ and $t_f$ are the near plane and the far plane of the sampling boundary. NeRF is obtained by optimizing the following loss function:

\begin{equation}
    \mathcal{L}_{\text {color}}=\sum_{\boldsymbol{r} \in \mathcal{R}}\left\|C(\boldsymbol{r})-\hat{C}(\boldsymbol{r})\right\|^{2},
\label{eq.color}
\end{equation}
where $\mathcal{R}$ represents all rays emitted from the pixels of training images.

\paragraph{DreamFusion.} A recent work DreamFusion~\cite{poole2022dreamfusion} shows that under the guidance of a 2D diffusion model $\phi$, a 3D implicit object represented by the NeRF $g_{\theta}$ can be generated from scratch according to a text prompt. In their method, the NeRF model $g_{\theta}$ renders an image $I$ at a random viewing angle, the pre-trained diffusion model $\phi$ sample noise $\epsilon$ at time-step $t$ to generate noisy image $I_{t} = I + \epsilon$. The main contribution of DreamFusion~\cite{poole2022dreamfusion} is that they proposed a gradient calculation by Score Distillation Sampling (SDS) loss~\cite{poole2022dreamfusion} to guide the update direction of NeRF:

\begin{equation}
\nabla_{\theta}\mathcal{L}_{ \mathrm {SDS}}(\phi,g_{\theta}) = \mathbb{E}_{t,\epsilon} \left[w(t) \left (\epsilon_{\phi}\left(I_t;y, t\right)-\epsilon\right) \frac{\partial I}{\partial\theta}\right],
\label{eq.SDS}
\end{equation}
where $w(t)$ is a weighted function correspond to time-step $t$, $y$ is a text embedding, $\epsilon_{\phi}$ is a learned denoising function. $\nabla_{\theta}\mathcal{L}_{\mathrm{SDS}}$ is used to update the NeRF network $g_{\theta}$ instead of propagating to the diffusion model $\phi$.  

\subsection{Semantic Mask Extraction}
\begin{figure}[H]
\centering
\subfigure[fortress scene \textbf{without} depth supervision.]{
\label{fortress_wo_depth}
\includegraphics[width=1\columnwidth]{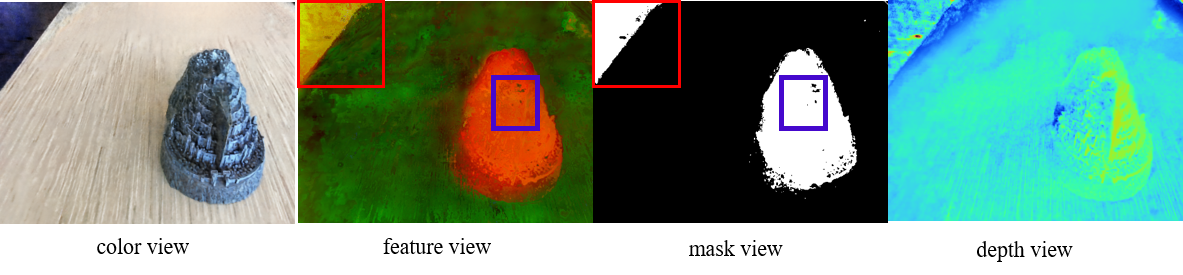}
}
\subfigure[fortress scene \textbf{with} depth supervision.]{
\label{fortress_wi_depth}
\includegraphics[width=1\columnwidth]{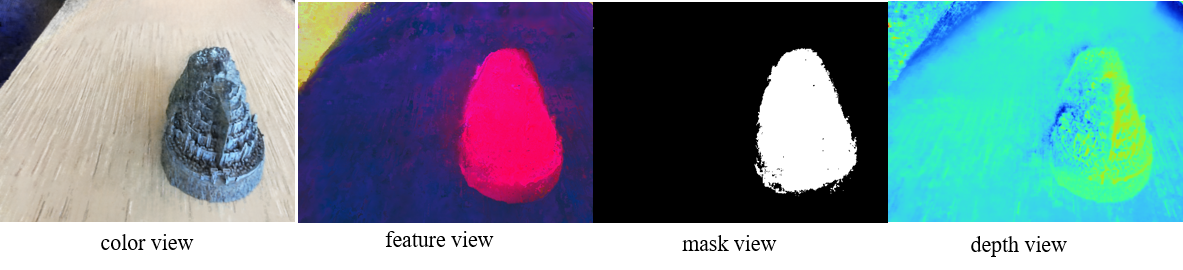}
}
\caption{\textbf{Ablation studies of depth supervision.}}
\label{Fig.fortress_depth}
\end{figure}

In our approach, the first step in modifying a 3D scene is to mask the target regions. Manually labeling each point in the training data or 3D field is very time-consuming. Obtaining the semantic information of a single two-dimensional image directly will lose the view consistency. Therefore, to extract accurate and consistent semantic information, we encode the feature $\boldsymbol{f}$ of view-independent point $\boldsymbol{x}$ in three-dimensional space. Note that the feature $\boldsymbol{f}$ is a high-dimensional vector with semantic information, which is different from explicit semantic labels in semantic segmentation. Here we follow DFF~\cite{kobayashi2022decomposing}. We first utilize an existing pre-trained model, such as CLIP~\cite{radford2021learning}, to extract the feature $\boldsymbol{f}$ of each training data. We additionally trained a feature network $s: \boldsymbol{x} \to \boldsymbol{f}$ and finally obtained the features from different views but keep view consistency through volume rendering equation:

\begin{equation}
    F(\boldsymbol{r})=\int_{t_{n}}^{t_{f}} T(t) \sigma(\boldsymbol{r}(t)) f(\boldsymbol{r}(t), d) \mathrm{d}t.   
\end{equation}
Similar to Eq.~\ref{eq.color}, we defined the loss function for optimizing the feature module:

\begin{equation}
\mathcal{L}_{\text {feature}}=\sum_{\boldsymbol{r} \in \mathcal{R}}\left\|F(\boldsymbol{r})-\hat{F}(\boldsymbol{r})\right\|^{2}.
\end{equation}
However, in some views, there will always be some noise affecting NeRF's depth estimation, resulting in imprecise semantic mask segmentation (see Fig.~\ref{Fig.fortress_depth}). Thus, we add depth information that is predicted by a pre-trained model~\cite{Ranftl2022} as a coarse depth supervision to mitigate this problem. At the same time, this measure also speeds up the convergence of the feature field:

\begin{equation}
\mathcal{L}_{\text {depth}}=\sum_{\boldsymbol{r} \in \mathcal{R}}\left\|D(\boldsymbol{r})-\hat{D}(\boldsymbol{r})\right\|^{2},
\end{equation}
where $D(\boldsymbol{r})$ is the depth predicted by a pre-trained model as the depth ground truth, and $\hat{D}(\boldsymbol{r})$ is the depth predicted by NeRF. Therefore, the final loss function for the first stage of our method is:

\begin{equation}
    \mathcal{L}_{\text {first-stage}} =  \mathcal{L}_{\text {color}} +  \lambda_{\text{feature}}\mathcal{L}_{\text {feature}} + \lambda_\text{depth}\mathcal{L}_{\text {depth}}.
\label{eq.s1}
\end{equation}
We do not obtain the semantic mask by comparing the features $\boldsymbol{f}$ of each 3D point $\boldsymbol{x}$ with the target feature. Instead, we compare the patch features with pixel features after rendering the feature map $F_{I}$. We prefer to use patch features instead of text features because we find that a higher threshold can be set to control more accurate segmentation. The mask is obtained using the following equation:

\begin{equation}
I_\mathrm{{mask}}^{H\times W \times  1}
= \mathbb{1} (Sim(F_{\mathrm{patch}},F_{I}) > \alpha),
\end{equation}
where $F_\mathrm{{patch}}$ is the mean feature of a selected patch, $Sim$ is a similarity function, and $\alpha$ is a threshold.

\begin{figure*}[t]
    \centering
    \includegraphics[width=2\columnwidth]{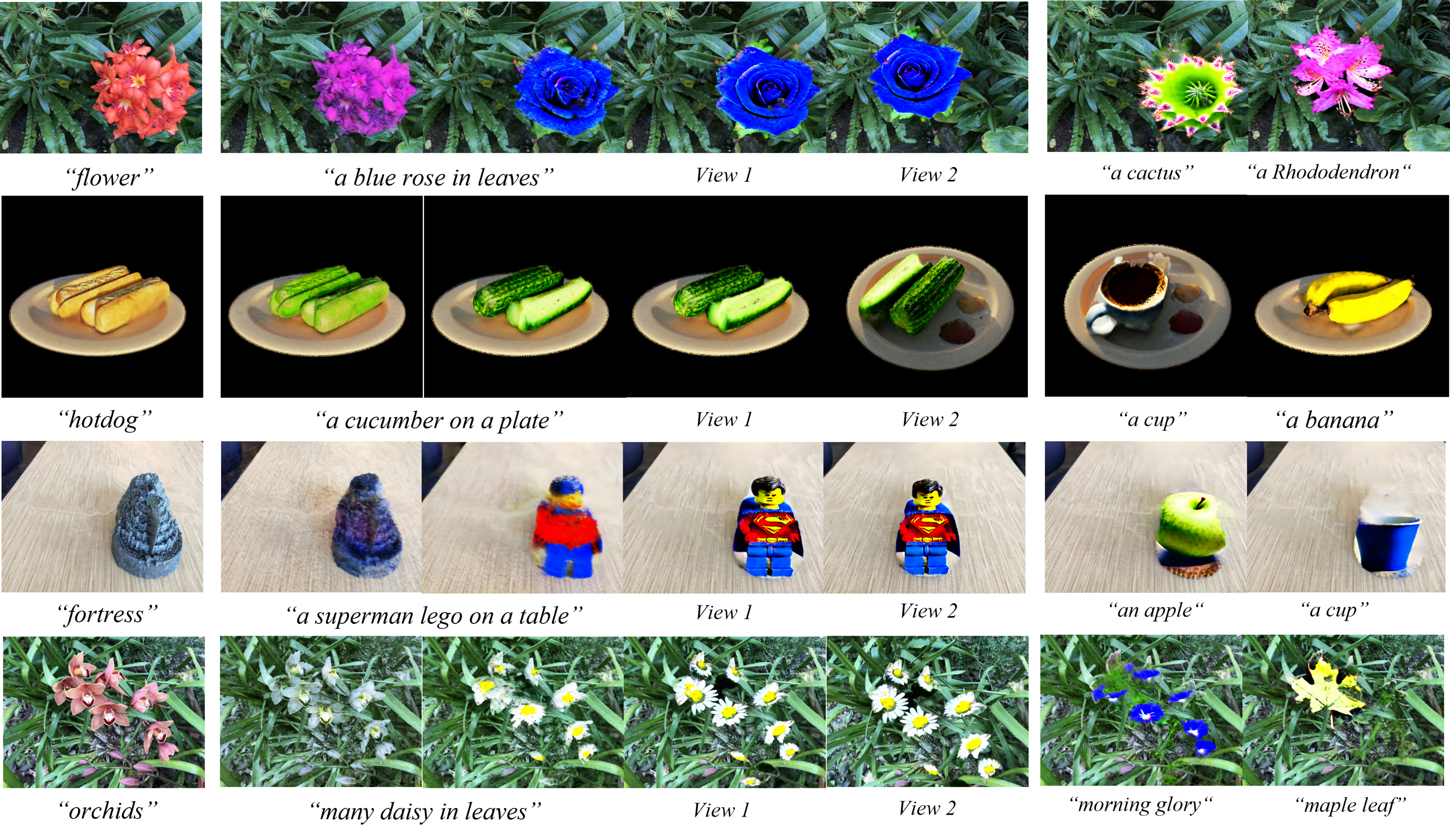}
    \caption{\textbf{Qualitative editing results.} We test our method on the Blender and LLFF. Our method can change the shape and appearance of objects in both real-world and synthetic-world datasets with a simple text prompt while maintaining almost high fidelity.}
    \label{Fig.result}
\end{figure*}

\begin{figure}
    \centering
    \includegraphics[width=1.0\columnwidth]{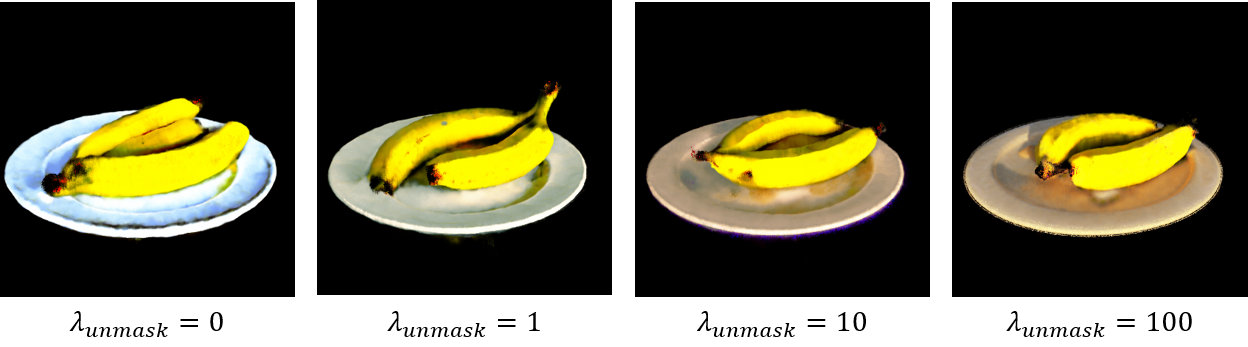}
    \caption{\textbf{Comparisons of different $\lambda_{\mathrm{{unmask}}}$.} The text prompt and background prompt we use here are \textit{``a banana in a plate"} and \textit{``an empty plate"}.}
    \label{Fig.lambda}
\end{figure}

\subsection{Text-to-3D Content Editing}
In the second stage, based on DreamFusion~\cite{poole2022dreamfusion} and the previously extracted masks, we modify the 3D content inside it on a pre-trained NeRF. Our goal is to keep the surrounding content unchanged while editing NeRF content. Thus, we keep optimizing the unmasked region by minimizing:

\begin{equation}
\mathcal{L}_{\mathrm{unmask}}=\sum_{\boldsymbol{r} \in \mathcal{R}}\left\|C(\boldsymbol{r})_{\mathrm{unmask}}-\hat{C}(\boldsymbol{r})_{\mathrm{unmask}}\right\|^{2},
\label{eq.unmask}
\end{equation}
\begin{equation}
    \mathrm{where} \quad C(\boldsymbol{r})_{\text{unmask}} = C(\boldsymbol{r}) \times \mathbb{1}(I_\text{{mask}} < 0.5),
\end{equation}
\begin{equation}
    \quad \hat{C}(\boldsymbol{r})_{\text{unmask}} = \hat{C}(\boldsymbol{r}) \times  \mathbb{1}(I_{\text{mask}} < 0.5).
\end{equation}
However, we find that when the diffusion model~\cite{rombach2022high} is used to guide the masked part to be modified, the newly generated small target object will expose the part covered by the previous object, which is largely unseen. Thus, we add a background prompt (BGT) that could partially solve this issue. For example, we use prompt \textit{``a blue rose in leaves"} instead of \textit{``a blue rose"}, so the BGT here is \textit{``leaves"}. (see the first-row example in Fig.~\ref{Fig.result}). Inspired by~\cite{weder2022removing}, we also add a CLIP loss function to guide the masked region for generating a background that could fill the black hole around the mask edge. The CLIP loss function is defined as:

\begin{equation}
    \mathcal{L}_\mathrm{{clip}} = - Sim\left(Z_I, Z_{\mathrm{BGT}}\right),
\label{eq.clip}
\end{equation}
where $Z_{I}$ and $Z_{\mathrm{BGT}}$ are the latent feature of rendering image $I$ and background prompt encoded by pre-trained CLIP model~\cite{li2022language}.

The insight of our method is that keep the diffusion model watch the whole scene by using $\nabla_{\theta}\mathcal{L}_{ \mathrm {SDS}}$ to guide the update direction of NeRF $g_{\theta}$ for optimizing the target region, and using $\mathcal{L}_{\mathrm{unmask}}$ to ensure the unmasked region of the target keep stable. Besides, the background prompt and CLIP loss function $\mathcal{L}_\mathrm{{clip}}$ is to make the unseen region to be filled.

Now the final loss function of our second stage is defined as:

\begin{equation}
     \mathcal{L}_\mathrm{{repaint}} = \lambda_\mathrm{{unmask}}\mathcal{L}_\mathrm{{unmask}} + \lambda_\mathrm{{clip}}\mathcal{L}_\mathrm{{clip}},
\label{eq.s2}
\end{equation}
where the $\lambda_\mathrm{unmask}$ and $\lambda_\mathrm{{clip}}$ is set here for balancing the change of the target object and other regions.

\subsection{Implementation Details}
Our implementation is divided into two parts, based on DFF~\cite{kobayashi2022decomposing} and DreamFusion~\cite{poole2022dreamfusion} separately. Each training iteration needs to render a whole view, this undoubtedly consumes a huge amount of memory, and the training speed is also crucial to user experience. Thus, we use Instant-NGP~\cite{muller2022instant} as our NeRF model, which is based on a multi-resolution hash grid structure to accelerate the training and rendering process. We test our method on a single NVIDIA RTX 3090 GPU. It is worth noting that the actual generation time depends on the appearance and shape of editing before and after. For example, if you only change the color of one car, it might only take 5 minutes. But turning a car into a chocolate candy car can take more than 40 minutes to achieve decent results. Please refer to our supplementary material for more details.

\paragraph{Mask Extraction.} In the first stage, we first extract the semantic masks of the training images. CLIP~\cite{radford2021learning} is a text-image pair-based multimodal model for self-supervised training, which is trained for aligning text information and image information. However, the feature maps extracted by CLIP are not at the pixel level, so here we use the LSeg~\cite{li2022language} to extract the feature maps as our training set. LSeg~\cite{li2022language} is a semantic segmentation model that is trained based on the CLIP weights. All feature maps are interpolated to the size of image size $H \times W \times 512$. We visualize the high dimensional feature via PCA, which can reflect the distribution of semantic information to a certain extent. In addition, for depth information, we use MiDAS~\cite{Ranftl2022}, a robust monocular depth estimation model, to extract rough depth information. The size of each depth image is $H \times W \times 1$, and the depth information is normalized to a range of $0$ to $1$. Thanks to Instant-NGP~\cite{muller2022instant} and the depth information we added as supervision, we only need to train 2000 steps for each scene to get a clear semantic mask. We use Adam~\cite{kingma2014adam} to optimize our NeRF model in the first stage, with a learning rate 1e-2 and batch size 4096.

\paragraph{3D Object Editing.} In the second stage, we only use the color images and the masks extracted in the first stage as inputs. We first train a NeRF by optimizing Eq.~\ref{eq.color} as our base model. In the pre-training phase, we sample a whole image in each iteration, so that we can observe the memory usage and adjust the hyperparameters in time to make compromises for the pre-trained diffusion model and CLIP model in the generation phase. For example, in the fortress scene, we set the bound to 1.4 and the learning rate to 1e-3 to ensure that the GPU does not exceed the video memory space as much as possible. For the training of generation, each scene will require an input of an object text prompt with a background text prompt. We use Stable Diffusion~\cite{rombach2022high} to supervise the update direction of the NeRF model and use the CLIP~\cite{radford2021learning} model to compare the similarity between the unmasked image (an image that includes the surrounding content and a masked blacked hole) and the BGT. We train these two phases using Adan with a learning rate of 1e-3 decaying to 1e-4 and a batch size of 1. These two phases are optimized for 3000 steps and 10000 steps respectively, more steps will be added for better visual effect.

\section{Experiments}
In this section, we focus on evaluating our method on different scenes with different prompts, we show qualitative editing results on Local Light Field Fusion (LLFF)~\cite{mildenhall2019llff} and Blender, followed by comparison experiments and ablation studies.

\begin{figure}
    \centering
    \subfigure[Mask + CLIP]{
        \includegraphics[width=0.9\columnwidth]{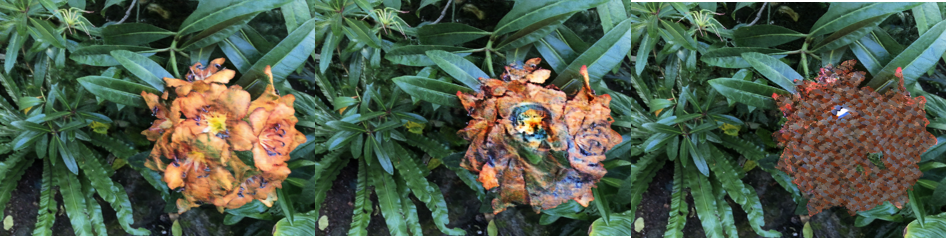}
        \label{Mask+CLIP}
    }
        \subfigure[Mask + SDS]{
        \includegraphics[width=0.9\columnwidth]{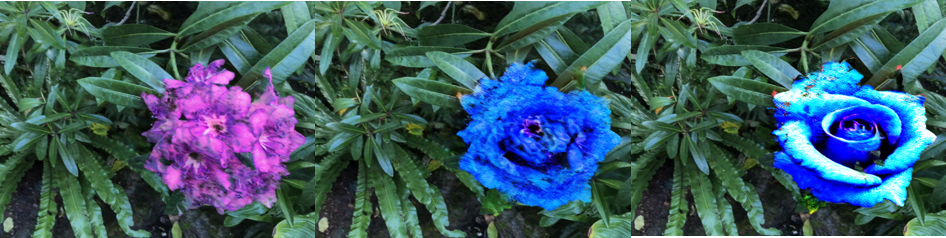}
        \label{Mask+CLIP}
    }
    \caption{We compared the ability of generations under two different guidance. Here we use text prompt \textit{``a blue rose"} in both tests. }
    \label{Fig.clip_sds}
\end{figure}

\subsection{Datasets}
We use Local Light Field Fusion (LLFF)~\cite{mildenhall2019llff} and Blender for testing. The LLFF~\cite{mildenhall2019llff} is collected from the real world in the form of shooting forward-facing, and its capture resolution is $4032 \times 3024$. The Blender comes from the synthetic world by rendering on 3D models, its resolution is  $800 \times 800$. In order to save the memory of the GPU, we resize the image size to $504 \times 378$ and the image size of Blender to $400 \times 400$. Our experiments show that our method is very effective for object editing in both worlds.

\subsection{Semantic Mask Extraction}

The role of the first part is mainly to replace the operation of manual masking and provides view-consistent masks for the second part. Thus, we mainly focus on how to get a clean and accurate mask view faster. In the Fig.~\ref{Fig.fortress_depth}, we take the fortress scene as an example, and our goal is to extract the objects of the fortress. The result shows that training for only 2000 steps without depth supervision produces inaccurate and noisy semantic masks. However, in our proposed model with depth informative supervision, 2000-step training yields more accurate and clean semantic masks. 

\begin{figure}[H]
\centering  
\includegraphics[width=1\columnwidth]{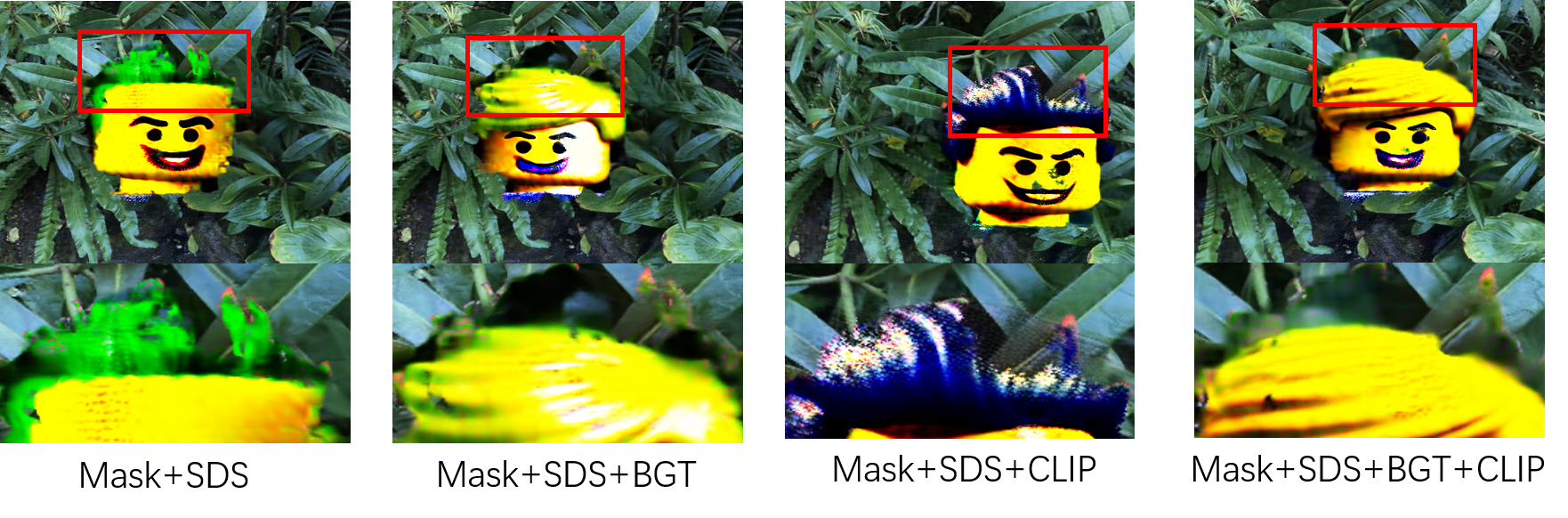}
\caption{\textbf{Ablation studies of background prompt guidance.} For \textbf{Mask+SDS}, the text prompt is \textit{``a lego man head"}; For \textbf{Mask+SDS+BGT}, the text prompt is \textit{``a lego man head in leaves"}; For \textbf{Mask+SDS+CLIP}, the text prompt is \textit{``a lego man"}, and the prompt for CLIP is \textit{``leaves"}; For \textbf{Mask+SDS+BGT+CLIP}, the text prompt is \textit{``a lego man head in leaves"}, and prompt for CLIP is \textit{``leaves"}.}
\label{Fig.bg}
\end{figure}

\subsection{Text-to-3D Editing}

We have conducted a lot of experiments on LLFF~\cite{mildenhall2019llff} and Blender, as shown in Fig.~\ref{Fig.result}, our method can recreate a photo-realistic target area while ensuring that the other original scene does not change or changes slightly. The first column of Fig.~\ref{Fig.result} is the view rendered by pre-trained NeRF. The text under the view of the first column is the target object we desire to change. Columns 2, 3, and 4 represent the process of gradually modifying the object. The text prompt is set below those three views, which includes the target prompt and the BGT. The BGT we send for the CLIP model is to encourage background generation. Columns 4 and 5 are different views that are rendered by modified NeRF. Columns 6 and 7 contain our other editing results under the same scene, the text below there is the target generation prompt, and the background prompts are consistent with the previous ones. We find that the generation of similar shapes works better and faster, such as turning a flower into a blue rose in the first row and a hotdog into a cucumber in the second row. For shapes that are not too similar, the generated effect will decrease, but it will gradually approach the content of the prompt, such as turning a hotdog into a cup of coffee in the first row and a fortress into an apple in the third row. At the same time, in our method, the model can perceive the surrounding objects and generate areas that have not been seen before. For example, in the showcase of the fortress to an apple, the edge of the fortress that previously covered part becomes material and color similar to the desktop. In the example of orchids to maple leaves, except for the modified maple leaf, other content that was orchids before has become the content of the grass in the background.

\paragraph{Comparisons of CLIP and SDS.} We compare the ability of Stable Diffusion~\cite{rombach2022high} and CLIP~\cite{radford2021learning} to guide object changes under the same mask. Fig.~\ref{Fig.clip_sds} shows the editing effect of the two models during the optimization process. The target text prompt here we set is \textit{``a blue rose"}, and the red flower is the source object. We first set Eq.~\ref{eq.unmask} to ensure that the surrounding content remains unchanged. For \textbf{Mask+CLIP}, we make sure that the CLIP model~\cite{radford2021learning} will only see the region of the flower and we train the NeRF model by optimizing the similarity loss function of Eq.~\ref{eq.clip}. We find that the CLIP model is weak to guide the NeRF model to the target direction very well. At the beginning of the training, the CLIP model is trying to make the flower move to a little bit blue, while after several epochs, the flower becomes an unreasonable texture. For \textbf{Mask+SDS}, we set the SDS loss~\cite{poole2022dreamfusion} to guide the masked part to become a blue rose. The result of \textbf{Mask+SDS} in the Fig.~\ref{Fig.clip_sds} shows that the model generates an excellent texture and shape. The comparison shows that the \textbf{Mask+SDS} can produce a much better effect than \textbf{Mask+CLIP}. 

\paragraph{Ablation studies of background prompt guidance.} We perform background-cued ablation experiments (see Fig.~\ref{Fig.bg}). For \textbf{Mask+SDS}, we only set an SDS loss~\cite{poole2022dreamfusion} to guide the masked region with an object text prompt, here we set is \textit{``a lego man head"}. The result shows that the model generates a shape like green hair on the Lego head, and the generated shape of the hair is still the shape of the petal. For \textbf{Mask+SDS+BGT}, here we set the text prompt for the diffusion model as \textit{``a lego man head in leaves"}, which means we explicitly tell the diffusion model what is the content around the Lego man. And the result is that hair of the Lego man is more reasonable, but the previously covered part on the top of the hair becomes a black hole. For \textbf{Mask+SDS+CLIP}, we remove the background prompt in the text prompt and provide a background prompt to CLIP, we are inspired by~\cite{mirzaei2022laterf} here. We give a view of the unmasked region to CLIP to fill the black hole. The result of this way is that the Lego man owns reasonable hair but with the shape of a petal. For the final \textbf{Mask+SDS+BGT+CLIP}, we set the text prompt includes the background prompt, which is \textit{``a lego man head in leaves"}. And the CLIP will receive a view of the unmasked image and a background prompt \textit{``leaves"}. The final results show that the hair becomes a normal shape, and the black hole is filled with a kind of green material. We also could see the result of \textbf{Mask+SDS+BGT+CLIP} in Fig.~\ref{Fig.result}, for example, some green leaves appeared around the blue rose in the example of flower to a blue rose.

\paragraph{Effects of Different Mask Weight.} In addition, we compare the effect of different $\lambda_{\mathrm{unmask}}$ in Eq.~\ref{eq.s2} (see Fig.~\ref{Fig.lambda}). The Stable Diffusion~\cite{poole2022dreamfusion} model we use in our method receives full-resolution images without any mask. Thus, we set a loss function Eq.~\ref{eq.unmask} to strongly control the unmasked part to avoid unwanted content. We use the hotdog scene as an example, and the goal is to change the hotdog to a banana. In the first experiment, we set $\lambda_{\mathrm{unmask}}=0$, which means no constraints on the background part (i.e., the plate), and the result is the diffusion model totally changes the shape and appearance of the plate. Then we gradually increase the weight $\lambda_{\mathrm{unmask}}$ to control the shape of the plate, the result shows that $\lambda_{\mathrm{unmask}}=100$ could retain the original shape and appearance of the plate well. Thus, in the practice of all scenes, we usually set $\lambda_{\mathrm{unmask}}=100$, we also find a weight too larger will extremely influence the speed of generation, which means more constraint to the ability of the diffusion model.

\begin{figure}
    \centering
    \includegraphics[width=1\columnwidth]{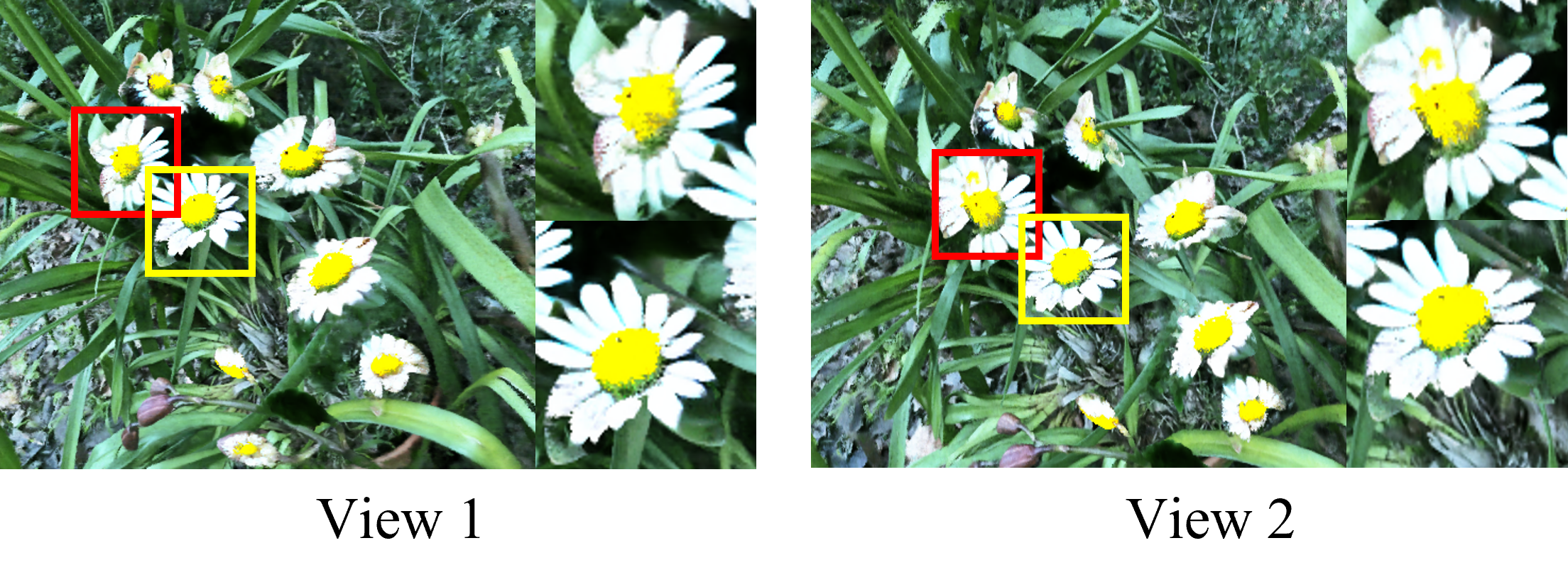}
    \caption{The daisy in the \textbf{red} frame is composed of two previous orchids; the daisy in the \textbf{yellow} frame is normal.}
    \label{fig.lim}
\end{figure}

\section{Limitations}
Although our method can modify the content of NeRF through text prompts, there are still some limitations. The first and most obvious problem with our method is that the whole process is time-consuming and space-consuming. For this reason, we cannot extend to a larger resolution of data or a larger scene. Several recent works~\cite{lin2022magic3d,metzer2022latent} have introduced how to solve the time problems of DreamFusion~\cite{poole2022dreamfusion}, but it still takes more than 30 minutes to generate a relatively high-quality 3D model at the object level. In another aspect, we find that in some cases, especially the new object shape is hugely different from the old one, the training process will be very tough for the model (see Fig.~\ref{Fig.result}, the cup from the hotdog is not completely changed). The last limitation is that angle constraint will influence the shape of the generation. For example, in Fig.~\ref{fig.lim}, the generated daisy on the edge of the view is actually combined by two orchids, while the generated daisy in the central location is much more normal.

\section{Conclusion}

In this paper, we proposed a NeRF editing framework based on the pre-trained 2D diffusion model guidance. Our method mainly obtained the edited scene by regenerating the masked part through the gradient guidance of the diffusion model. Moreover, we added a background prompt and a CLIP loss to ease the problem of invisible background. At the same time, we also added depth information as supervision in the semantic mask acquisition part of the task for a faster training speed and better semantic masks. Our method combined the semantic information of the scene and used a text-driven method to modify the content of the scene. Compared with explicitly modifying the 3D scene, our method would make the NeRF more diverse and promote the development of the simulation environment based on neural rendering. We believe that there will be faster and better editing methods in the future to further improve the editability of NeRF models.

\clearpage 
\section*{Acknowledgements}
This work was supported in part by the National Natural Science Foundation of China (NSFC) under Grant 62002238, and the Open Research Fund from Guangdong Laboratory of Artificial Intelligence and Digital Economy (SZ) under Grant GML-KF-22-26.

\bibliographystyle{named}
\bibliography{repaint-nerf}

\end{document}